\documentclass[runningheads]{llncs}
\usepackage{graphicx}
\usepackage{comment}
\usepackage{amsmath,amssymb} 
\usepackage{color}
\usepackage{subfigure}
\usepackage{multirow}
\usepackage{rotating}
\usepackage[pagebackref=true,breaklinks=true,letterpaper=true,colorlinks,bookmarks=false,citecolor=blue,linkcolor=blue]{hyperref}

\begin{document}
\pagestyle{headings}
\mainmatter
\def\ECCVSubNumber{2581}  

\title{Multi-Scale Positive Sample Refinement for Few-Shot Object Detection} 

\renewcommand{\thefootnote}{*}
\titlerunning{Multi-Scale Positive Sample Refinement for Few-Shot Object Detection}
%
\author{Jiaxi Wu\inst{1,2,3} \and
Songtao Liu\inst{1,2,3} \and
Di Huang\inst{1,2,3}\thanks{indicates corresponding author (ORCID: 0000-0002-2412-9330).} \and
Yunhong Wang\inst{1,3}}
\authorrunning{J. Wu, S. Liu, D. Huang, Y. Wang}
%
\institute{BAIC for BDBC, Beihang University, Beijing 100191, China\\ \and
SKLSDE, Beihang University, Beijing 100191, China\\ \and
SCSE, Beihang University, Beijing 100191, China\\
\email{\{wujiaxi,liusongtao,dhuang,yhwang\}@buaa.edu.cn}}
\maketitle

\begin{abstract}
Few-shot object detection (FSOD) helps detectors adapt to unseen classes with few training instances, and is useful when manual annotation is time-consuming or data acquisition is limited.
Unlike previous attempts that exploit few-shot classification techniques to facilitate FSOD, this work highlights the necessity of handling the problem of scale variations, which is challenging due to the unique sample distribution.
To this end, we propose a Multi-scale Positive Sample Refinement (MPSR) approach to enrich object scales in FSOD. 
It generates multi-scale positive samples as object pyramids and refines the prediction at various scales. 
We demonstrate its advantage by integrating it as an auxiliary branch to the popular architecture of Faster R-CNN with FPN, delivering a strong FSOD solution. 
Several experiments are conducted on PASCAL VOC and MS COCO, and the proposed approach achieves state of the art results and significantly  outperforms other counterparts, which shows its effectiveness.
Code is available at \href{https://github.com/jiaxi-wu/MPSR}{https://github.com/jiaxi-wu/MPSR}.
\end{abstract}

\section{Introduction}

Object detection makes great progress these years following the success of deep convolutional neural networks (CNN) \cite{imagenet,resnet,fastrcnn,fasterrcnn,cascade}. 
These CNN based detectors generally require large amounts of annotated data to learn extensive numbers of parameters, and their performance significantly drops when training data are inadequate.
Unfortunately, for object detection, labeling data is quite expensive and the samples of some object categories are even hard to collect, such as endangered animals or tumor lesions. 
This triggers considerable attentions to effective detectors dealing with limited training samples.
Few-shot learning is a popular and promising direction to address this issue.
However, the overwhelming majority of the existing few-shot investigations focus on object/image classification, while the efforts on the more challenging \emph{few-shot object detection} (FSOD) task are relatively rare.

With the massive parameters of CNN models, training detectors from scratch with scarce annotations generally incurs a high risk of overfitting.
Preliminary research \cite{lstd} tackles this problem in a transfer learning paradigm. 
Given a set of base classes with sufficient annotations and some novel classes with only a few samples, the goal is to acquire meta-level knowledge from base classes and then apply it to facilitating few-shot learning in detection of novel classes.
Subsequent works~\cite{repmet,matchdet,yolore,metarcnn} strengthen this pipeline by bringing more advanced methods on few-shot image classification, and commonly emphasize to improve classification performance of  Region-of-Interest (RoI) in FSOD by using metric learning techniques. 
With elaborately learned representations, they ameliorate the similarity measurement between RoIs and marginally annotated instances, reporting better detection results.
Meanwhile,~\cite{yolore,metarcnn} also attempt to deliver more general detectors, which account for all the classes rather than the novel ones only, by jointly using their samples in the training phase.
\begin{figure}
	\centering
	\subfigure[Bus]{
		\includegraphics[width=5.6cm]{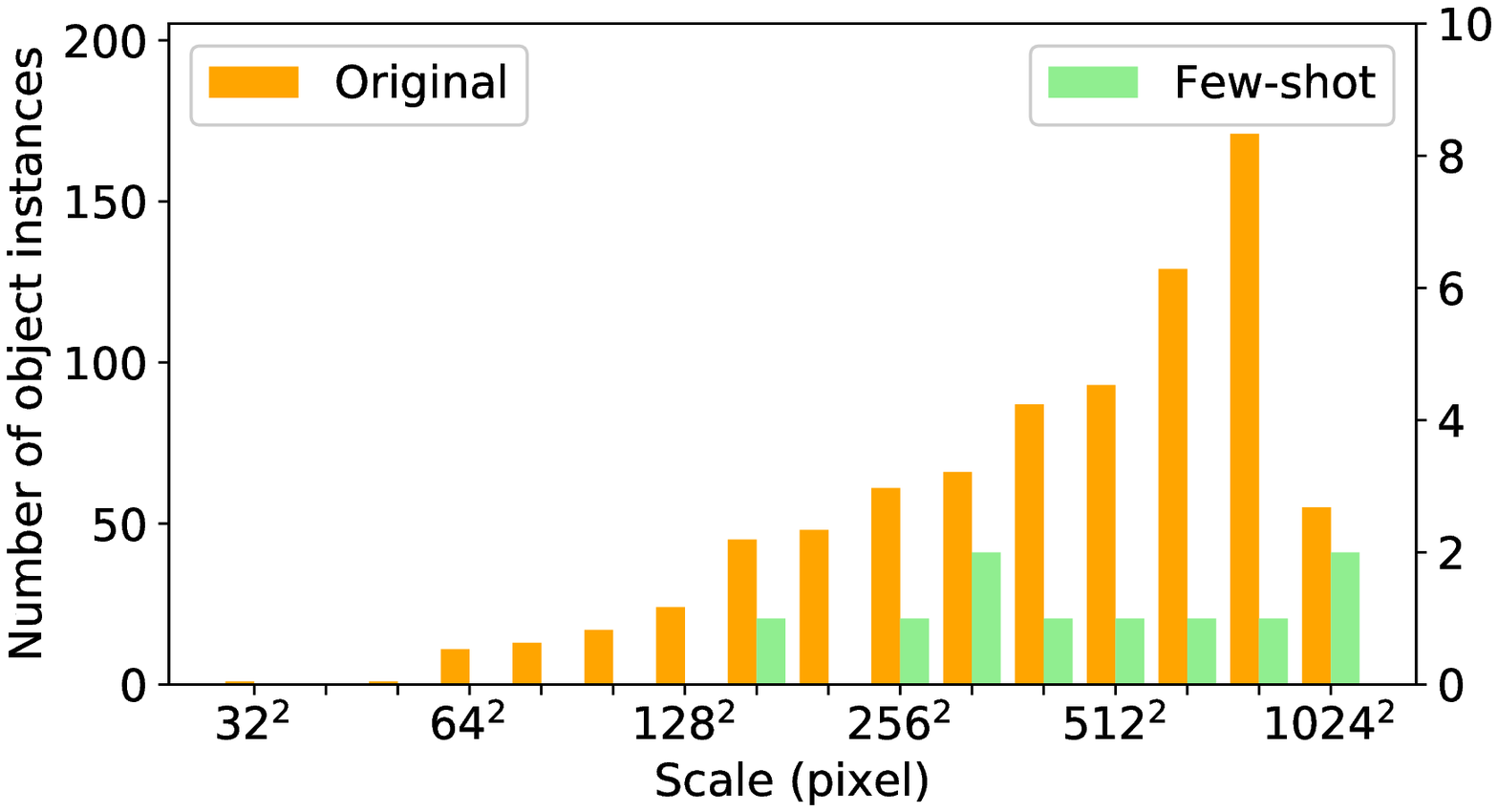}
	}
	\quad
	\subfigure[Cow]{
		\includegraphics[width=5.6cm]{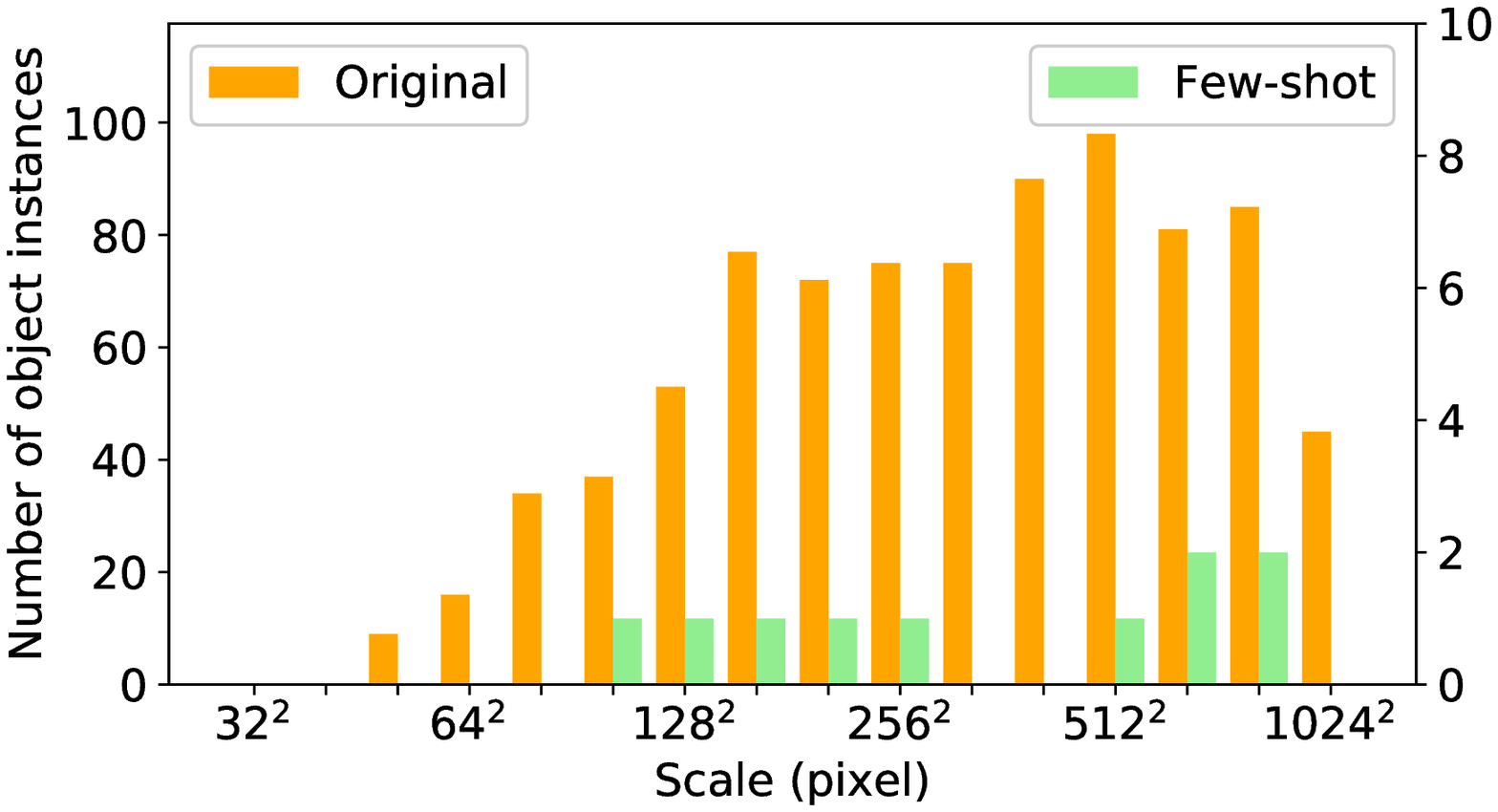}
	}
	\quad
	\caption{Illustration of scale distributions of two specific classes: (a) bus and (b) cow, in PASCAL VOC (Original) and a 10-shot subset (Few-shot). Images are resized with the shorter size at 800 pixels for statistics}
	\label{fig:sds}
\end{figure}

The prior studies demonstrate that the FSOD problem can be alleviated in a similar manner as few-shot image classification. 
Nevertheless, object detection is much more difficult than image classification, as it involves not only classification but also localization, where the threat of varying scales of objects is particularly evident.
The scale invariance has been widely explored in generic supervised detectors \cite{fastrcnn,snip,sniper,san}, while it remains largely intact in FSOD.
Moreover, restricted by the quantity of annotations, this scale issue is even more tricky.
As shown in Fig.~\ref{fig:sds}, the lack of labels of novel classes leads to a sparse scale space (green bars) which may be totally divergent from the original distribution (yellow bars) of abundant training data. 
One could assume to make use of current effective solutions from generic object detection to enrich the scale space.
For instance, Feature Pyramid Network (FPN), which builds multi-scale feature maps to detect objects at different scales, applies to situations where significant scale variations exist~\cite{fpn}.
This universal property does contribute to FSOD, but it will not mitigate the difference of the scale distribution in the data of novel classes.
Regarding image pyramids~\cite{sppnet,fastrcnn}, they build multi-scale representations of an image and allow detectors to capture objects in it at different scales.
Although they are expected to narrow such a gap between the two scale distributions, the case is not so straightforward.
Specifically, multi-scale inputs result in an increase in improper negative samples due to anchor matching.
These improper negative samples contain a part of features belonging to the positive samples, which interferes their recognition. 
With abundant data, the network learns to extract diverse contexts and suppress the improper local patterns. 
But it is harmful to FSOD where both semantic and scale distributions are sparse and biased.

In this work, we propose a Multi-scale Positive Sample Refinement (MPSR) approach to few-shot object detection, aiming at solving its unique challenge of sparse scale distribution.
We take the reputed Faster R-CNN as the basic detection model and employ FPN in the backbone network to improve its tolerance to scale variations. 
We then exploit an auxiliary refinement branch to generate multi-scale positive samples as object pyramids and further refine the prediction. 
This additional branch shares the same weights with the original Faster R-CNN.
During training, this branch classifies the extracted object pyramids in both the Region Proposal Network (RPN) and the detector head.
To keep scale-consistent prediction without introducing more improper negatives, we abandon the anchor matching rules and adaptively assign the FPN stage and spatial locations to the object pyramids as positives. It is worth noting that as we use no extra weights in training, our method achieves remarkable performance gains in an inference cost-free manner and can be conveniently deployed on different detectors.

The contributions of this study are three-fold:
\begin{enumerate}
\item To the best of our knowledge, it is the first work to discuss the scale problem in FSOD. We reveal the sparsity of scale distributions in FSOD with both quantitative and qualitative analysis.
\item To address this problem, we propose the MPSR approach to enrich the scale space without largely increasing improper negatives.
\item Comprehensive experiments are carried out, and significant improvements from MPSR demonstrate its advantage.
\end{enumerate}

\section{Related Work}

\subsubsection{Few-Shot Image Classification.}
There are relatively many historical studies in the area of few-shot image classification that targets recognition of objects with only a handful of images in each class \cite{oneshot,metanet}. 
\cite{maml} learns to initialize weights that effectively adapt to unseen categories. 
\cite{learnet,ppa} aim to predict network parameters without heavily training on novel images. 
\cite{siamese,matchnet,compare} employ metric learning to replace linear classifiers with learnable metrics for comparison between query and support samples. Although few-shot image classification techniques are usually used to advance the phase of RoI classification in FSOD, they are different tasks, as FSOD has to consider localization in addition.

\subsubsection{Generic Object Detection.}
Recent object detection architectures are mainly divided into two categories: one-stage detectors and two-stage detectors. 
One-stage detectors use a single CNN to directly predict bounding boxes \cite{yolo,yolov2,ssd,rfbnet}, and two-stage ones first generate region proposals and then classify them for decision making \cite{rcnn,fastrcnn,fasterrcnn}. 
Apart from network design, scale invariance is an important aspect to detectors and many solutions have recently been proposed to handle scale changes \cite{fpn,snip,sniper,san}. 
For example, \cite{fpn} builds multi-scale feature maps to match objects at different scales. 
\cite{snip} performs scale normalization to detect scale-specific objects and adopts image pyramids for multi-scale detection. 
These studies generally adapt to alleviate large size differences of objects. 
Few-shot object detection suffers from scale variations in a more serious way where a few samples sparsely distribute in the scale space. 

\subsubsection{Object Detection with Limited Annotations.}
To relieve heavy annotation dependence in object detection, there exist two main directions without using external data. 
One is weakly-supervised object detection, where only image-level labels are provided and spatial supervision is unknown \cite{wsddn}. 
Research basically concentrates on how to rank and classify region proposals with only coarse labels through multiple instance learning \cite{oicr,wsrpn,cmil}. 
Another is semi-supervised object detection that assumes abundant images are available while the number of bounding box annotations is limited \cite{video}. 
In this case, previous studies confirm the effectiveness of adopting extra images by pseudo label mining \cite{fewexample,notercnn} or multiple instance learning \cite{disease}. 
Both the directions reduce manual annotation demanding to some extent, but they heavily depend on the amount of training images. 
They have the difficulty in dealing with constrained conditions where data acquisition is inadequate, \emph{i.e.}, few-shot object detection.

\subsubsection{Few-Shot Object Detection.}
Preliminary work \cite{lstd} on FSOD introduces a general transfer learning framework and presents the Low-Shot Transfer Detector (LSTD), which reduces overfitting by adapting pre-trained detectors to few-shot scenarios with limited training images. 
Following this framework, RepMet \cite{repmet} incorporates a distance metric learning classifier into the RoI classification head in the detector. 
Instead of categorizing objects with fully-connected layers, RepMet extracts representative embedding vectors by clustering and calculates distances between query and annotated instances. 
\cite{matchdet} is motivated by \cite{siamese} which scores the similarity in a siamese network and computes pair-wise object relationship in both the RPN and the detection head. 
\cite{yolore} is a single-stage detector combined with a meta-model that re-weights the importance of features from the base model. 
The meta-model encodes class-specific features from annotated images at a proper scale, and the features are viewed as reweighting coefficients and fed to the base model. 
Similarly, \cite{metarcnn} delivers a two-stage detection architecture and re-weights RoI features in the detection head. 
Unlike previous studies where spatial influence is not considered, we argue that scale invariance is a challenging issue to FSOD, as the samples are few and their scale distribution is sparse.
We improve the detector by refining object crops rather than masked images \cite{yolore,metarcnn} or siamese inputs \cite{matchdet} for additional training, which enriches the scale space and ensures the detector being fully trained at all scales.

\section{Background}
Before introducing MPSR, we briefly review the standard protocols and the basic detector we adopt for completeness. As it is the first work that addresses the challenge of sparse scale distribution in FSOD, we conduct some preliminary attempts with the current effective methods from generic object detection (\emph{i.e.}, FPN and image pyramids) to enrich the scale space and discuss their limitations.

\subsection{Baseline Few-Shot Object Detection}

\subsubsection{Few-Shot Object Detection Protocols.}
Following the settings in \cite{yolore,metarcnn}, object classes are divided into base classes with abundant data and novel classes with only a few training samples.
The training process of FSOD generally adopts a two-step paradigm. During base training, the detection network is trained with a large-scale dataset that only contains base classes. 
Then the detection network is fine-tuned on the few-shot dataset, which only contains a very small number of balanced  training samples for both base and novel classes.
This two-step training schedule avoids the risk of overfitting with insufficient training samples on novel classes. 
It also prevents the detector from extremely imbalanced training if all annotations from both base and novel classes are exploited together \cite{metarcnn}.
To build the balanced few-shot dataset, \cite{yolore} employs the $k$-shot sampling strategy, where each object class only has $k$ annotated bounding boxes.
Another work \cite{lstd} collects $k$ images for each class in the few-shot dataset. 
As $k$ images actually contain an arbitrary number of instances, training and evaluation under this protocol tend to be unstable. 
We thus use the former strategy following \cite{yolore}.

\subsubsection{Basic Detection Model.}
With the fast development in generic object detection, the base detector in FSOD has many choices.
\cite{yolore} is based on YOLOv2 \cite{yolov2}, which is a single-stage 
detector.
\cite{metarcnn} is based on a classical two-stage detector, Faster R-CNN \cite{fasterrcnn}, and demonstrates that Faster R-CNN provides consistently better results.
Therefore, we take the latter as our basic detection model.
Faster R-CNN consists of the RPN and the detection head.
For a given image, the RPN head generates proposals with objectness scores and bounding-box regression offsets. 
The RPN loss function is:
\begin{align}
L_{RPN} = \frac{1}{N_{obj}}{\sum_{i = 1}^{N_{obj}}L_{Bcls}^{i}} + \frac{1}{N_{obj}}{\sum_{i = 1}^{N_{obj}}L_{Preg}^{i}}.
\end{align}
For the $i$th anchor in a mini-batch, $L_{Bcls}^{i}$ is the binary cross-entropy loss over background and foreground and $L_{Preg}^{i}$ is the smooth $L_{1}$ loss defined in \cite{fasterrcnn}. $N_{obj}$ is the total number of chosen anchors. 
These proposals are used to extract RoI features and then fed to the detection (RoI) head that outputs class-specific scores and bounding-box regression offsets. 
The loss function is defined as:
\begin{align}
L_{RoI} = \frac{1}{N_{RoI}}{\sum_{i = 1}^{N_{RoI}}L_{Kcls}^{i}} + \frac{1}{N_{RoI}}{\sum_{i = 1}^{N_{RoI}}L_{Rreg}^{i}},
\end{align}
where $L_{Kcls}^{i}$ is the log loss over $K$ classes and $N_{RoI}$ is the number of RoIs in a mini-batch. 
Different from the original implementation in \cite{fasterrcnn}, we employ a class-agnostic regression task in the detection head, which is the same as \cite{lstd}.
The total loss is the sum of $L_{RPN}$ and $L_{RoI}$.

\subsection{Preliminary Attempts}

\subsubsection{FPN for Multi-Scale Detection.}
As FPN is commonly adopted in generic object detection to address the scale variation issue \cite{fpn,cascade}, we first consider applying it to FSOD in our preliminary experiments. 
FPN generates several different semantic feature maps at different scales, enriching the scale space in features.  
Our experiments validate that it is still practically useful under the restricted conditions in FSOD. 
We thus exploit Faster R-CNN with FPN as our second baseline.
However, FPN does not change the distribution in the data of novel classes and the sparsity of scale distribution remains unsolved in FSOD.

\begin{figure}
	\centering
	\includegraphics[height=4cm]{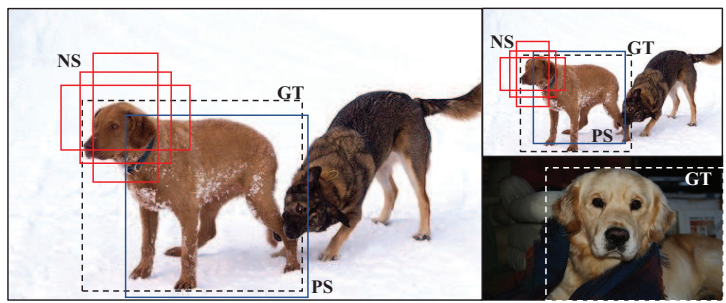}
	\caption{An example of improper negative samples in FSOD. Negative samples (NS), positive samples (PS) and ground-truth (GT) bounding boxes are annotated. The improper negative samples significantly increase as more scales are involved (top right), while they may even be true positives in other contexts (bottom right)}
	\label{fig:neg}
\end{figure}

\subsubsection{Image Pyramids for Multi-Scale Training.} 
To enrich object scales, we then consider a multi-scale training strategy which is also widely used in generic object detection for multi-scale feature extraction \cite{sppnet,fastrcnn} or data augmentation \cite{yolov2}. 
In few-shot object detection, image pyramids enrich object scales as data augmentation and the sparse scale distribution can be theoretically solved.
However, this multi-scale training strategy acts differently in FSOD with the increasing number of improper negative samples.
As in Fig.~\ref{fig:neg}, red bounding boxes are negative samples in training while they actually contain part of objects and may even be true positive samples in other contexts (as in bottom right).
These improper negative samples require sufficient contexts and clues to suppress, inhibiting being mistaken for potential objects. 
Such an interference is trivial when abundant annotations are available, but it is quite harmful to the sparse and biased distribution in FSOD.
Moreover, with multi-scale training, a large number of extra improper negative samples are introduced, which further hurts the performance.

\section{Multi-Scale Positive Sample Refinement}


\subsection{Multi-Scale Positive Sample Refinement Branch}
Motivated by the above discussion, we employ FPN in the backbone of Faster R-CNN as the advanced version of baseline.
To enrich scales of positive samples without largely increasing improper negative samples, we extract each object independently and resize them to various scales, denoted as object pyramids.
Specifically, each object is cropped by a square window (whose side is equal to the longer side of the bounding box) with a minor random shift. It is then resized to $\left\{ 32^{2},64^{2},128^{2},256^{2},512^{2},800^{2} \right\}$ pixels, which is similar to anchor design. 
\begin{figure}
	\centering
	\includegraphics[height=4.5cm]{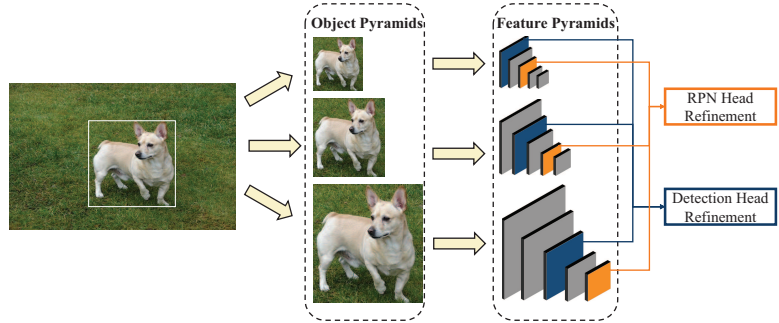}
	\caption{Multi-scale positive sample feature extraction. The positive sample is extracted and resized to various scales. Specific feature maps from FPN are selected for refinement}
	\label{fig:scales}
\end{figure}

In object pyramids, each image only contains a single instance, which is inconsistent to the standard detection pipeline.
Therefore, we propose an extra positive sample refinement branch to adaptively project the object pyramids into the standard detection network.
For a given object, the standard FPN pipeline samples the certain scale level and the spatial locations as positives for training, operated by anchor matching.
However, performing anchor matching on cropped single objects is wasteful and also incurs more improper negatives that hurt the performance for FSOD.
As shown in Fig.~\ref{fig:scales}, instead of anchor matching, we manually select the corresponding scale level of feature maps and the fixed center locations as positives for each object, keeping it consistent with the standard FPN assigning rules.
After selecting specific features from these feature maps, we feed them directly to the RPN head and the detection head for refinement.
\setlength{\tabcolsep}{4pt}
\begin{table}
	\begin{center}
		\caption{FPN feature map selection for different object scales. For each object, two specific feature maps are activated, fed to RPN and detection (RoI) heads respectively}
		\label{table:scalereftable}
		\begin{tabular}{c|cccccc}
			\hline
			& $32^{2}$ & $64^{2}$ & $128^{2}$ & $256^{2}$ & $512^{2}$ & $800^{2}$ \\
			\hline
			RPN & $P_{2}$ & $P_{3}$ & $P_{4}$  & $P_{5}$  & $P_{6}$  & $P_{6}$  \\
			RoI & $P_{2}$ & $P_{2}$ & $P_{2}$  & $P_{3}$  & $P_{4}$  & $P_{5}$  \\
			\hline
		\end{tabular}
	\end{center}
\end{table}
\setlength{\tabcolsep}{1.4pt}

In the RPN head, the multi-scale feature maps of FPN  $\left\{ P_{2},P_{3},P_{4},P_{5},P_{6} \right\}$ represent anchors whose areas are $\left\{ 32^{2},64^{2},128^{2},256^{2},~512^{2} \right\}$ pixels respectively. 
For a given object, only one feature map with the consistent scale is activated, as shown in Table~\ref{table:scalereftable}.
To simulate that each proposal is predicted by its center location in RPN, we select centric $2^{2}$ features for object refinement.
We also put anchors with $\left\{ 1:2,1:1,2:1 \right\}$ aspect ratios on the sampled locations.  
These selected anchors are viewed as positives for the RPN classifier. 

To extract RoI features for the detection head, only $\left\{ P_{2},P_{3},P_{4},P_{5} \right\}$ are used and the original RoI area partitions in the standard FPN pipeline are: $\left( {0^{2},112^{2}} \right)$, $\left\lbrack 112^{2},224^{2}~ \right)$, $\left\lbrack 224^{2},448^{2}~ \right)$, $\left\lbrack 448^{2},\infty~ \right)$~\cite{fpn}. 
We also select one feature map at a specific scale for each object to keep the scale consistency, as shown in Table~\ref{table:scalereftable}. 
As the randomly cropped objects tend to have larger sizes than the orignal ground truth bounding boxes, we slightly increase the scale range of each FPN stage for better selection.
Selected feature maps are adaptively pooled to the same RoI size and fed to the RoI classifier. 

\subsection{Framework}
\begin{figure}
	\centering
	\includegraphics[height=7cm]{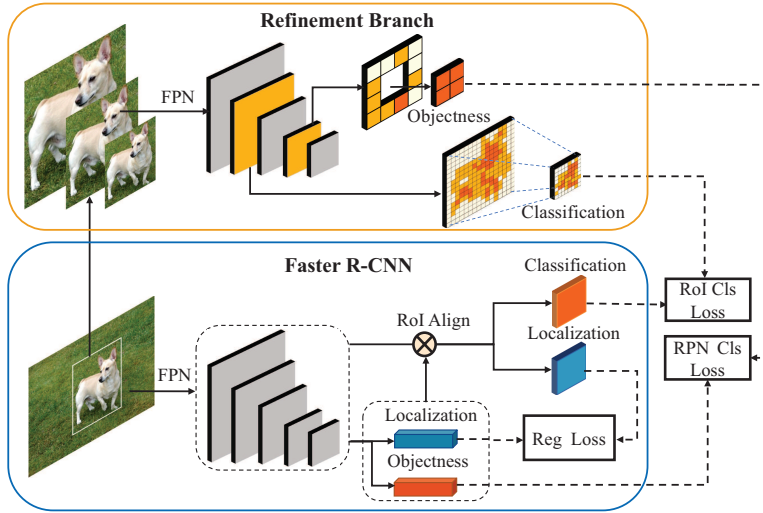}
	\caption{MPSR architecture. On an input image to Faster R-CNN, the auxiliary branch extracts samples and resizes them to different scales. Each sample is fed to the FPN and specific features are selected to refine RPN and RoI heads in Faster R-CNN}
	\label{fig:arch}
\end{figure}

As shown in Fig.~\ref{fig:arch}, the whole detection framework for training consists of Faster R-CNN with FPN and the refinement branch working in parallel while sharing the same weights.
For a given image, it is processed by the backbone network, RPN, RoI Align layer, and the detection head in the standard two-stage detection pipeline \cite{fasterrcnn}. 
Simultaneously, an independent object extracted from the original image is resized to different scales as object pyramids. 
The object pyramids are fed into the detection network as described above. 
The outputs from RPN and detection heads in the MPSR branch include objectness scores and class-specific scores similar to the definitions in Section 3.1.
The loss function of the RPN head containing Faster R-CNN and the MPSR branch is defined as:
\begin{align}
L_{RPN} = \frac{1}{N_{obj}{+ M}_{obj}}{\sum_{i = 1}^{N_{obj}{+ M}_{obj}}L_{Bcls}^{i}} + \frac{1}{N_{obj}}{\sum_{i = 1}^{N_{obj}}L_{Preg}^{i}},
\end{align}
where $M_{obj}$ is the number of selected positive anchor samples for refinement. The loss function of the detection head is defined as:
\begin{align}
L_{RoI} = \frac{1}{N_{RoI}}{\sum_{i = 1}^{N_{RoI}}L_{Kcls}^{i}} + \frac{\lambda}{M_{RoI}}{\sum_{i = 1}^{M_{RoI}}L_{Kcls}^{i}} + \frac{1}{N_{RoI}}{\sum_{i = 1}^{N_{RoI}}L_{Rreg}^{i}},
\end{align}
where $M_{RoI}$ is the number of selected RoIs in MPSR. 
Unlike the RPN head loss where $M_{obj}$ is close to $N_{obj}$, the number of positives from object pyramids is quite small compared to $N_{RoI}$ in the RoI head.
We thus add a weight parameter $\lambda$ to the RoI classification loss of the positives from MPSR to adjust its magnitude, which is set to 0.1 by default.
After the whole network is fully trained, the extra MPSR branch is removed and only Faster R-CNN with FPN is used for inference. 
Therefore, the MPSR approach that we propose benefits FSOD training without extra time cost at inference.

\section{Experiments}

\subsection{Datasets and Settings}
We evaluate our method on the PASCAL VOC 2007 \cite{voc07}, 2012 \cite{voc12} and MS COCO \cite{coco} benchmarks. 
For fair quantitative comparison with state of the art (SOTA) methods, we follow the setups in \cite{yolore,metarcnn} to construct few-shot detection datasets.

\subsubsection{PASCAL VOC.}
Our networks are trained on the modified VOC 2007 trainval and VOC 2012 trainval sets. 
The standard VOC 2007 test set is used for evaluation. 
The evaluation metric is the mean Average Precision (mAP). 
Both the trainval sets are split by object categories, where 5 are randomly chosen as novel classes and the left 15 are base classes. 
Here we follow \cite{yolore} to use the same three class splits, where the unseen classes are \{``bird'', ``bus'', ``cow'', ``motorbike'' (``mbike''), ``sofa''\}, \{``aeroplane'' (``aero''), ``bottle'', ``cow'', ``horse'', ``sofa''\}, \{``boat'', ``cat'', ``motorbike'', ``sheep'', ``sofa''\}, respectively. 
For FSOD experiments, the few-shot dataset consists of images where only $k$ object instances are available for each category and $k$ is set as 1/3/5/10. 

\subsubsection{MS COCO.}
COCO has 80 object categories, where the 20 categories overlapped with PASCAL VOC are denoted as novel classes. 
5,000 images from the val set, denoted as minival, are used for evaluation while the left images in the train and val sets are used for training. 
Base and few-shot dataset construction is the same as that in PASCAL VOC except that $k$ is set as 10/30. 

\subsubsection{Implementation Details.}
We train and test detection networks on images of a single scale. 
We resize input images so that their shorter sides are set to $800$ pixels and the longer sides are less than 1,333 pixels while maintaining the aspect ratio. 
Our backbone is ResNet-101 \cite{resnet} with the RoI Align~\cite{maskrcnn} layer and we use the weights pre-trained on ImageNet~\cite{imagenet} in initialization.
For efficient training, we randomly sample one object to generate the object pyramid for each image. 
After training on base classes, only the last fully-connected layer (for classification) of the detection head is replaced. 
The new classification layer is randomly initialized and none of the network layers is frozen during few-shot fine-tuning. 
We train our networks with a batchsize of 4 on 2 GPUs, 2 images per GPU. We run the SGD optimizer with the momentum of 0.9 and the parameter decay of 0.0001. 
For base training on VOC, models are trained for 240k, 8k, and 4k iterations with learning rates of 0.005, 0.0005 and 0.00005 respectively. 
For few-shot fine-tuning on VOC, we train models for 1,300, 400, 300 iterations and the learning rates are 0.005, 0.0005 and 0.00005, respectively. 
Models are trained on base COCO classes for 56k, 14k, and 10k iterations.
For COCO few-shot fine-tuning, the 10-shot dataset requires 2,800, 700, and 500 iterations, while the 30-shot dataset requires 5,600, 1,400, 1,000 iterations.

\subsection{Results}

We compare our results with two baseline methods (denoted as Baseline and Baseline-FPN) as well as two SOTA few-shot detection counterparts. 
Baseline and Baseline-FPN are our implemented Faster R-CNN and Faster R-CNN with FPN described in Section 3.
YOLO-FS \cite{yolore} and Meta R-CNN \cite{metarcnn} are the SOTA few-shot detectors based on DarkNet-19 and ResNet-101, respectively.
It should be noted that due to better implementation and training strategy, our baseline achieves higher performance than SOTA, which is also confirmed by the very recent work~\cite{ftfsod}.

\setlength{\tabcolsep}{1.7pt}
\begin{table}
	\begin{center}
		\caption{Comparison of different methods in terms of mAP (\%) of novel classes using the three splits on the VOC 2007 test set}
		\label{table:mainvoctable}
		\begin{tabular}{c|cccc|cccc|cccc}
			\hline
			& \multicolumn{4}{c|}{Class Split 1}                            & \multicolumn{4}{c|}{Class Split 2}                            & \multicolumn{4}{c}{Class Split 3}                             \\ \hline
			Method/Shot  & 1             & 3             & 5             & 10            & 1             & 3             & 5             & 10            & 1             & 3             & 5             & 10            \\ \hline
			YOLO-FS~\cite{yolore}      & 14.8          & 26.7          & 33.9          & 47.2          & 15.7          & 22.7          & 30.1          & 39.2          & 19.2          & 25.7          & 40.6          & 41.3          \\
			Meta R-CNN~\cite{metarcnn}   & 19.9          & 35.0          & 45.7          & 51.5          & 10.4          & 29.6          & 34.8          & 45.4          & 14.3          & 27.5          & 41.2          & 48.1          \\ \hline
			Baseline     & 24.5          & 40.8          & 44.6          & 47.9          & 16.7          & 34.9          & 37.0          & 40.9          & 27.3          & 36.3          & 41.2          & 45.2          \\
			Baseline-FPN & 25.5          & 41.1          & 49.6          & 56.9          & 15.5          & 37.7          & 38.9          & 43.8          & 29.9          & 37.9          & 46.3          & 47.8          \\ \hline
			MPSR (ours)  & \textbf{41.7} & \textbf{51.4} & \textbf{55.2} & \textbf{61.8} & \textbf{24.4} & \textbf{39.2} & \textbf{39.9} & \textbf{47.8} & \textbf{35.6} & \textbf{42.3} & \textbf{48.0} & \textbf{49.7} \\ \hline
		\end{tabular}
	\end{center}
\end{table}
\setlength{\tabcolsep}{1.4pt}
\subsubsection{PASCAL VOC.}
 MPSR achieves 82.1\%/82.7\%/82.9\% on base classes of three splits respectively before few-shot fine-tuning.
The main results of few-shot experiments on VOC are summarized in Table~\ref{table:mainvoctable}.
It can be seen from this table that the results of the two baselines (\emph{i.e.} Baseline and Baseline-FPN) are close to each other when the number of instances is extremely small (\emph{e.g.} 1 or 3), and Baseline-FPN largely outperforms the other as the number of images increases. 
This demonstrates that FPN benefits few-shot object detection as in generic object detection. 
Moreover, our method further improves the performance of Baseline-FPN with any number of training samples in all the three class splits.
Specifically, by solving the sparsity of object scales, we achieve a significant increase in mAP compared to the best scores of the two baselines, particularly when training samples are extremely scarce, \emph{e.g.} 16.2\% on 1-shot split-1. 
It clearly highlights the effectiveness of the extra MPSR branch. Regarding other counterparts \cite{yolore,metarcnn}, the proposed approach outperforms them by a large margin, reporting the state of the art scores on this dataset.

\setlength{\tabcolsep}{2pt}
\begin{table}
	\begin{center}
		\caption{AP (\%) of each novel class on the 3-/10-shot VOC dataset of the first class split. mAP (\%) of novel classes and base classes are also presented}
		\label{table:detailedvoctable}
		\begin{tabular}{cc|ccccc|cc}
			\hline
			&            & \multicolumn{5}{c|}{Novel Classes}                                            & \multicolumn{2}{c}{Mean}     \\ \hline
			\multicolumn{1}{c|}{Shot}        & Method            & bird          & bus           & cow           & mbike         & sofa          & Novel         & Base          \\ \hline
			\multicolumn{1}{c|}{\multirow{5}{*}{3}}  & YOLO-FS~\cite{yolore}      & 26.1          & 19.1          & 40.7          & 20.4          & 27.1          & 26.7          & 64.8          \\
			\multicolumn{1}{c|}{}                    & Meta R-CNN\cite{metarcnn}   & 30.1          & 44.6          & 50.8          & 38.8          & 10.7          & 35.0          & 64.8          \\
			\multicolumn{1}{c|}{}                    & Baseline     & 34.9          & 26.9          & 53.3          & 50.8          & 38.2          & 40.8          & 45.2          \\
			\multicolumn{1}{c|}{}                    & Baseline-FPN & 32.6          & 29.4          & 45.5          & 56.2          & 41.7          & 41.1          & 66.2          \\
			\multicolumn{1}{c|}{}                    & MPSR (ours)  & \textbf{35.1} & \textbf{60.6} & \textbf{56.6} & \textbf{61.5} & \textbf{43.4} & \textbf{51.4} & \textbf{67.8} \\ \hline
			\multicolumn{1}{c|}{\multirow{5}{*}{10}} & YOLO-FS~\cite{yolore}      & 30.0          & 62.7          & 43.2          & 60.6          & 39.6          & 47.2          & 63.6          \\
			\multicolumn{1}{c|}{}                    & Meta R-CNN~\cite{metarcnn}   & \textbf{52.5} & 55.9          & 52.7          & 54.6          & 41.6          & 51.5          & 67.9          \\
			\multicolumn{1}{c|}{}                    & Baseline     & 38.6          & 48.6          & 51.6          & 57.2          & 43.4          & 47.9          & 47.8          \\
			\multicolumn{1}{c|}{}                    & Baseline-FPN & 41.8          & 68.4          & 61.7          & 66.8          & 45.8          & 56.9          & 70.0          \\
			\multicolumn{1}{c|}{}                    & MPSR (ours)  & 48.3          & \textbf{73.7} & \textbf{68.2} & \textbf{70.8} & \textbf{48.2} & \textbf{61.8} & \textbf{71.8} \\ \hline
		\end{tabular}
	\end{center}
\end{table}
\setlength{\tabcolsep}{1.4pt}

Following~\cite{yolore,metarcnn}, we display the detailed results of 3-/10-shot detection in the first split on VOC in Table~\ref{table:detailedvoctable}. 
Consistently, our Baseline-FPN outperforms the existing methods on both the novel and base classes. 
This confirms that FPN addresses the scale problem in FSOD to some extent. 
Furthermore, our method improves the accuracies of Baseline-FPN in all the settings by integrating MPSR, illustrating its advantage.

\subsubsection{MS COCO.}
We evaluate the method using 10-/30-shot setups on MS COCO with the standard COCO metrics. The results on novel classes are provided in Table~\ref{table:cocotable}. 
Although COCO is quite challenging, we still achieve an increase of 0.4\% on 30-shot compared with Baseline-FPN while boosting the SOTA mAP from 12.4\% (Meta R-CNN) to 14.1\%.
Specifically, our method improves the recognition of small, medium and large objects simultaneously. 
This demonstrates that our balanced scales of input objects are effective.
\setlength{\tabcolsep}{0.2pt}
\begin{table}
\begin{center}
\caption{AP (\%) and AR (\%) of 10-/30-shot scores of novel classes on COCO minival}
\label{table:cocotable}
\begin{tabular}{c|c|ccc|ccc|ccc|ccc}
	\hline
	Shot                & Method       & AP            & $AP_{50}$          & $AP_{75}$          & $AP_{S}$          & $AP_{M}$           & $AP_{L}$           & $AR_{1}$           & $AR_{10}$          & $AR_{100}$         & $AR_{S}$          & $AR_{M}$           & $AR_{L}$           \\ \hline
	\multirow{5}{*}{10} & YOLO-FS~\cite{yolore}      & 5.6           & 12.3          & 4.6           & 0.9          & 3.5           & 10.5          & 10.1          & 14.3          & 14.4          & 1.5          & 8.4           & 28.2          \\
	& Meta R-CNN~\cite{metarcnn}   & 8.7           & \textbf{19.1} & 6.6           & 2.3          & 7.7           & 14.0          & 12.6          & 17.8          & 17.9          & \textbf{7.8} & 15.6          & 27.2          \\
	& Baseline     & 8.8           & 18.7          & 7.1           & 2.9          & 8.1           & 15.0          & 12.9          & 17.2          & 17.2          & 4.1          & 14.2          & 29.1          \\
	& Baseline-FPN & 9.5           & 17.3          & 9.4           & 2.7          & 8.4           & 15.9          & 14.8          & 20.6          & 20.6          & 4.7          & 19.3          & 33.1          \\
	& MPSR (ours)  & \textbf{9.8}  & 17.9          & \textbf{9.7}  & \textbf{3.3} & \textbf{9.2}  & \textbf{16.1} & \textbf{15.7} & \textbf{21.2} & \textbf{21.2} & 4.6          & \textbf{19.6} & \textbf{34.3} \\ \hline
	\multirow{5}{*}{30} & YOLO-FS~\cite{yolore}      & 9.1           & 19.0          & 7.6           & 0.8          & 4.9           & 16.8          & 13.2          & 17.7          & 17.8          & 1.5          & 10.4          & 33.5          \\
	& Meta R-CNN~\cite{metarcnn}   & 12.4          & 25.3          & 10.8          & 2.8          & 11.6          & 19.0          & 15.0          & 21.4          & 21.7          & \textbf{8.6} & 20.0          & 32.1          \\
	& Baseline     & 12.6          & \textbf{25.7} & 11.0          & 3.2          & 11.8          & 20.7          & 15.9          & 21.8          & 21.8          & 5.1          & 18.0          & 36.9          \\
	& Baseline-FPN & 13.7          & 25.1          & 13.3          & 3.6          & 12.5          & \textbf{23.3} & \textbf{17.8} & \textbf{24.7} & \textbf{24.7} & 5.4          & \textbf{21.6} & \textbf{40.5} \\
	& MPSR (ours)  & \textbf{14.1} & 25.4          & \textbf{14.2} & \textbf{4.0} & \textbf{12.9} & 23.0          & 17.7          & 24.2          & 24.3          & 5.5          & 21.0          & 39.3          \\ \hline
\end{tabular}
\end{center}
\end{table}
\setlength{\tabcolsep}{1.4pt}

\subsubsection{MS COCO to PASCAL VOC.}
We conduct cross-dataset experiments on the standard VOC 2007 test set. 
In this setup, all the models are trained on the base COCO dataset and finetured with 10-shot objects in novel classes on VOC.
Results of Baseline and Baseline-FPN are 38.5\% and 39.3\% respectively. 
They are worse than 10-shot results only trained on PASCAL VOC due to the large domain shift. 
Cross-dataset results of YOLO-FS and Meta R-CNN are 32.3\% and 37.4\% respectively. 
Our MPSR achieves 42.3\%, which indicates that our method has better generalization ability in cross-domain situations.

\subsection{Analysis of Sparse Scales}
We visualize the scale distribution of two categories on the original dataset (Pascal VOC) and 10-shot subset in Fig.~\ref{fig:sds}.
It is obvious that the scale distribution in the few-shot dataset is extremely sparse and distinct from the original ones.

\setlength{\tabcolsep}{4pt}
\begin{table}
	\begin{center}
		\caption{AP (\%) on bus/cow class. Two 10-shot datasets are constructed on VOC split-1, where scales of instances are random or limited. Std over 5 runs are presented}
		\label{table:extremetable}
		\begin{tabular}{c|cc|cc}
			\hline
			& \multicolumn{2}{c|}{Bus} & \multicolumn{2}{c}{Cow} \\ \hline
			Method       & Random     & Limited     & Random      & Limited     \\ \hline
			Baseline-FPN & 68.4$\pm$0.6       & 39.5$\pm$1.3        & 61.7$\pm$0.9        & 39.9$\pm$1.2         \\
			MPSR (ours)  & 73.7$\pm$1.6       & 54.0$\pm$1.4          & 68.2$\pm$1.0        & 52.5$\pm$1.6        \\ \hline
		\end{tabular}
	\end{center}
\end{table}
\setlength{\tabcolsep}{1.4pt}

To quantitatively analyze the negative effect of scale sparsity, we evaluate detectors on two specific 10-shot datasets. 
We carefully select the bus and cow instances with the scale between $128^2$ and $256^2$ pixels to construct the ``limited'' few-shot datasets. 
As shown in Table~\ref{table:extremetable}, such extremely sparse scales lead to a significant drop in performance (\emph{e.g.} for bus, -28.9\% on Baseline-FPN). 
Therefore, it is essential to solve the extremely sparse and biased scale distribution in FSOD.
With our MPSR, the reduction of performance is relieved.

\setlength{\tabcolsep}{5pt}
\begin{table}
	\begin{center}
		\caption{mAP (\%) comparison of novel/base classes on VOC split-1: Baseline-FPN, SNIPER~\cite{sniper},  Baseline-FPN with scale augmentation/image pyramids and MPSR}
		\label{table:iptable}
		\begin{tabular}{c|ccc|ccc}
			\hline
			\multicolumn{1}{c|}{} & \multicolumn{3}{c|}{Novel}                    & \multicolumn{3}{c}{Base}                      \\ \hline
			Method/Shot           & 1             & 3             & 5             & 1             & 3             & 5             \\ \hline
			Baseline-FPN          & 25.5          & 41.1          & 49.6          & 56.9          & 66.2          & 67.9          \\
			SNIPER~\cite{sniper} & 1.4          & 21.0          & 39.7          & \textbf{67.8}          & \textbf{74.8}          & \textbf{76.2}          \\
			Scale Augmentation        & 29.8          & 44.7          & 49.8          & 52.7          & 67.1          & 68.8          \\
			Image Pyramids        & 29.5          & 48.4          & 50.4          & 58.1          & 67.5          & 68.3          \\
			MPSR (ours)           & \textbf{41.7} & \textbf{51.4} & \textbf{55.2} & 59.4 & 67.8 & 68.4 \\ \hline
		\end{tabular}
	\end{center}
\end{table}
\setlength{\tabcolsep}{1.4pt}

As in Table~\ref{table:iptable}, we compare MPSR with several methods that are used for scale invariance.
SNIPER~\cite{sniper} shows a lower accuracy on novel classes and a higher accuracy on base classes than the baseline. 
As SNIPER strictly limits the scale range in training, it actually magnifies the sparsity of scales in FSOD. 
Such low performance also indicates the importance of enriching scales.  
We also evaluate the scale augmentation and image pyramids with a shorter side of $\{480,576,688,864,1200\}$ \cite{sppnet}.
We can see that our MPSR achieves better results than those two  multi-scale training methods on the novel classes. 
When only one instance is available for each object category, our method exceeds multi-scale training by $\sim$12\%, demonstrating its superiority.

\subsection{Ablation Studies}
We conduct some ablation studies to verify the effectiveness of the proposed manual selection and refinement method in Table~\ref{table:rpnroitable}.
\setlength{\tabcolsep}{5pt}
\begin{table}
	\begin{center}
		\caption{mAP (\%) of MPSR with different settings of novel classes on VOC split-1}
		\label{table:rpnroitable}
		\begin{tabular}{ccccc|ccc}
			\hline
			\multirow{2}{*}{\begin{tabular}[c]{@{}c@{}}Baseline\\ FPN\end{tabular}} &
			\multirow{2}{*}{\begin{tabular}[c]{@{}c@{}}Object\\ Pyramids\end{tabular}} & \multicolumn{1}{c|}{\multirow{2}{*}{\begin{tabular}[c]{@{}c@{}}Manual\\ Selection\end{tabular}}} & \multicolumn{2}{c|}{Refinement} & \multicolumn{3}{c}{Shot} \\ \cline{4-8} 
			& \multicolumn{2}{c|}{} & RPN & RoI & 1 & 3 & 5 \\ \hline
			$\checkmark$ & &  &  &  & 25.5 & 41.1 & 49.6 \\
			$\checkmark$ & $\checkmark$ &  & $\checkmark$ & $\checkmark$ & 30.8 & 43.6 & 49.6 \\
			$\checkmark$ & $\checkmark$ & $\checkmark$ & $\checkmark$ &  & 36.7 & 48.0 & 54.4 \\
			$\checkmark$ & $\checkmark$ & $\checkmark$ &  & $\checkmark$ & 33.7 & 48.2 & 54.7 \\
			$\checkmark$ & $\checkmark$ & $\checkmark$ & $\checkmark$ & $\checkmark$ & 41.7 & 51.4 & 55.2 \\ \hline
		\end{tabular}
	\end{center}
\end{table}
\setlength{\tabcolsep}{1.4pt}

\subsubsection{Manual Selection.}
From the first two lines in Table~\ref{table:rpnroitable}, we see that applying anchor matching to object pyramids on both RPN and RoI heads achieves better performance than Baseline-FPN. 
However, when compared to the last three lines with manual selection rules, anchor matching indeed limits the benefits of object pyramids, as it brings more improper negative samples to interfere few-shot training. 
It confirms the necessity of the proposed manual refinement rules. 

\subsubsection{RPN and Detection Refinement.}
As in the last three lines of Table~\ref{table:rpnroitable}, we individually evaluate RPN refinement and detection (RoI) refinement to analyze their credits in the entire approach.
Models with only the RPN and RoI refinement branches exceed Baseline-FPN in all the settings, which proves their effectiveness. 
Our method combines them and reaches the top score, which indicates that the two branches play complementary roles.

\section{Conclusions}

This paper targets the scale problem caused by the unique sample distribution in few-shot object detection. To deal with this issue, we propose a novel approach, namely multi-scale positive sample refinement. It generates multi-scale positive samples as object pyramids and refines the detectors at different scales, thus enlarging the scales of positive samples while limiting improper negative samples. We further deliver a strong FSOD solution by integrating MPSR to Faster R-CNN with FPN as an auxiliary branch. Experiments are extensively carried out on PASCAL VOC and MS COCO, and the proposed approach reports better scores compared to current state of the arts, which shows its advantage.

\section*{Acknowledgment}

This work is funded by the Research Program of State Key Laboratory of Software Development Environment (SKLSDE-2019ZX-03) and the Fundamental Research Funds for the Central Universities.

\clearpage
%
%
\bibliographystyle{splncs04}
\bibliography{egbib}
\clearpage
\appendix
\section{Appendix}
\subsection{Ablation of Refinement in the Two-Step Training}
By default, we apply the refinement branch to detectors during both training steps for consistency.
To reveal the effectiveness of our method in the two steps, we evaluate detectors with refinement only during base training or few-shot fine-tuning individually.
As shown in Table~\ref{table:baseandfewrefine}, refining detectors only during base training gets better results than Baseline-FPN, which means detection at various scales benefits from our method to some extent.
Refining detectors only during few-shot fine-tuning exceeds refining base only and Baseline-FPN by a large margin when the number of instances is extremely small (\emph{e.g.} 1 or 3), which demonstrates that our method relieves the unique sparsity of scales in FSOD.
Besides, refining detectors during both base training and few-shot fine-tuning achieves the best results, indicating that the two steps play complementary roles.

\setlength{\tabcolsep}{5pt}
\begin{table}
	\begin{center}
		\caption{mAP (\%) of novel classes on VOC split-1 applying refinement during different training stages}
		\label{table:baseandfewrefine}
		\begin{tabular}{c|cccc}
			\hline
			Method/Shot & 1 & 3 & 5 & 10 \\ \hline
			Baseline-FPN & 25.5 & 41.1 & 49.6 & 56.9 \\
			Only Base & 26.6 & 43.5 & 50.7 & 59.4 \\
			Only Few-shot & 36.5 & 47.3 & 50.6 & 59.0 \\
			Both Steps & \textbf{41.7} & \textbf{51.4} & \textbf{55.2} & \textbf{61.8} \\ \hline
		\end{tabular}
	\end{center}
\end{table}
\setlength{\tabcolsep}{1.4pt}

\subsection{Complete Results on PASCAL VOC}
As shown in Table~\ref{table:alldetailed}, we present the complete results on PASCAL VOC as in~\cite{metarcnn}. In this table, we also supply the 2-shot experimental results for consistency.

\setlength{\tabcolsep}{1.5pt}
\begin{sidewaystable}
	\begin{center}
		\caption{AP (\%) of each novel class on the few-shot VOC datasets. mAP (\%) of novel classes are also presented}
		\label{table:alldetailed}
		\begin{tabular}{c|c|cccccc|cccccc|cccccc}
			\hline
			&  & \multicolumn{6}{c|}{Class Split 1} & \multicolumn{6}{c|}{Class Split 2} & \multicolumn{6}{c}{Class Split 3} \\ \hline
			Shot & Method & bird & bus & cow & mbike & sofa & mean & aero & bottle & cow & horse & sofa & mean & boat & cat & mbike & sheep & sofa & mean \\ \hline
			\multirow{5}{*}{1} & YOLO-FS~\cite{yolore} & 13.5 & 10.6 & 31.5 & 13.8 & 4.3 & 14.8 & 11.8 & 9.1 & 15.6 & 23.7 & 18.2 & 15.7 & 10.8 & 44.0 & 17.8 & 18.1 & 5.3 & 19.2 \\
			& Meta R-CNN~\cite{metarcnn} & 6.1 & 32.8 & 15.0 & 35.4 & 0.2 & 19.9 & \textbf{23.9} & 0.8 & 23.6 & 3.1 & 0.7 & 10.4 & 0.6 & 31.1 & 28.9 & 11.0 & 0.1 & 14.3 \\
			& Baseline & 21.0 & 14.3 & 21.6 & 50.6 & 15.0 & 24.5 & 12.7 & 9.1 & 9.7 & \textbf{42.5} & 9.8 & 16.7 & 9.9 & 47.0 & 43.7 & 24.1 & 11.9 & 27.3 \\
			& Baseline-FPN & 25.2 & 9.2 & 22.1 & 52.1 & 18.8 & 25.5 & 20.7 & \textbf{9.4} & 19.4 & 13.1 & 15.0 & 15.5 & 11.4 & 41.6 & 42.7 & 35.9 & 17.8 & 29.9 \\
			& MPSR (ours) & \textbf{33.5} & \textbf{41.2} & \textbf{57.6} & \textbf{54.5} & \textbf{21.6} & \textbf{41.7} & 21.2 & 9.1 & \textbf{36.0} & 30.9 & \textbf{25.1} & \textbf{24.4} & \textbf{14.9} & \textbf{47.8} & \textbf{57.7} & \textbf{34.7} & \textbf{22.8} & \textbf{35.6} \\ \hline
			\multirow{5}{*}{2} & YOLO-FS~\cite{yolore} & 21.2 & 12.0 & 16.8 & 17.9 & 9.6 & 15.5 & 28.6 & 0.9 & 27.6 & 0.0 & 19.5 & 15.3 & 5.3 & 46.4 & 18.4 & 26.1 & 12.4 & 21.7 \\
			& Meta R-CNN~\cite{metarcnn} & 17.2 & \textbf{34.4} & 43.8 & 31.8 & 0.4 & 25.5 & 12.4 & 0.1 & 44.4 & \textbf{50.1} & 0.1 & 19.4 & 10.6 & 24.0 & 36.2 & 19.2 & 0.8 & 18.2 \\
			& Baseline & 36.5 & 10.6 & 39.5 & 55.2 & 26.3 & 33.6 & 31.9 & 9.1 & \textbf{45.5} & 18.3 & 22.8 & 25.5 & 6.8 & \textbf{49.7} & 52.6 & 35.7 & 22.9 & 33.5 \\
			& Baseline-FPN & 31.3 & 15.1 & 41.2 & 51.5 & 25..9 & 33.0 & \textbf{39.5} & 5.6 & 42.6 & 19.5 & 24.4 & 26.3 & 16.5 & 46.5 & \textbf{61.4} & 34.5 & 25.7 & 36.9 \\
			& MPSR (ours) & \textbf{38.2} & 28.6 & \textbf{56.5} & \textbf{57.3} & \textbf{32.0} & \textbf{42.5} & 36.5 & \textbf{9.1} & 45.1 & 21.6 & \textbf{34.2} & \textbf{29.3} & \textbf{17.9} & 49.6 & 59.2 & \textbf{49.2} & \textbf{32.9} & \textbf{41.8} \\ \hline
			\multirow{5}{*}{3} & YOLO-FS~\cite{yolore} & 26.1 & 19.1 & 40.7 & 20.4 & 27.1 & 26.7 & 29.4 & 4.6 & 34.9 & 6.8 & 37.9 & 22.7 & 11.2 & 39.8 & 20.9 & 23.7 & 33.0 & 25.7 \\
			& Meta R-CNN~\cite{metarcnn} & 30.1 & 44.6 & 50.8 & 38.8 & 10.7 & 35.0 & 25.2 & 0.1 & \textbf{50.7} & \textbf{53.2} & 18.8 & 29.6 & \textbf{16.3} & 39.7 & 32.6 & \textbf{38.8} & 10.3 & 27.5 \\
			& Baseline & 34.9 & 26.9 & 53.3 & 50.8 & 38.2 & 40.8 & 42.8 & 6.1 & 49.6 & 42.0 & 34.2 & 34.9 & 14.4 & 54.8 & 48.1 & 32.4 & 31.8 & 36.3 \\
			& Baseline-FPN & 32.6 & 29.4 & 45.5 & 56.2 & 41.7 & 41.1 & 48.7 & \textbf{9.7} & 46.3 & 42.4 & 41.4 & 37.7 & 10.7 & 48.1 & \textbf{57.3} & 31.9 & 41.3 & 37.9 \\
			& MPSR (ours) & \textbf{35.1} & \textbf{60.6} & \textbf{56.6} & \textbf{61.5} & \textbf{43.4} & \textbf{51.4} & \textbf{49.2} & 9.1 & 47.1 & 46.3 & \textbf{44.3} & \textbf{39.2} & 14.4 & \textbf{60.6} & 57.1 & 37.2 & \textbf{42.3} & \textbf{42.3} \\ \hline
			\multirow{5}{*}{5} & YOLO-FS~\cite{yolore} & 31.5 & 21.1 & 39.8 & 40.0 & 37.0 & 33.9 & 33.1 & 9.4 & 38.4 & 25.4 & 44.0 & 30.1 & 14.2 & \textbf{57.3} & 50.8 & 38.9 & 41.6 & 40.6 \\
			& Meta R-CNN~\cite{metarcnn} & 35.8 & 47.9 & 54.9 & 55.8 & 34.0 & 45.7 & 28.5 & 0.3 & \textbf{50.4} & \textbf{56.7} & 38.0 & 34.8 & 16.6 & 45.8 & 53.9 & 41.5 & 48.1 & 41.2 \\
			& Baseline & 36.9 & 30.7 & \textbf{59.9} & 57.4 & 37.8 & 44.6 & 46.7 & \textbf{11.9} & 41.5 & 46.0 & 39.0 & 37.0 & 17.9 & 53.3 & 54.6 & 42.8 & 37.4 & 41.2 \\
			& Baseline-FPN & \textbf{40.9} & 52.1 & 45.2 & 64.3 & 45.6 & 49.6 & \textbf{50.0} & 11.2 & 39.7 & 48.2 & 45.3 & 38.9 & 20.2 & 48.7 & 67.6 & 47.0 & \textbf{48.2} & 46.3 \\
			& MPSR (ours) & 39.7 & \textbf{65.5} & 55.1 & \textbf{68.5} & \textbf{47.4} & \textbf{55.2} & 47.8 & 10.4 & 45.2 & 47.5 & \textbf{48.8} & \textbf{39.9} & \textbf{20.9} & 56.6 & \textbf{68.1} & \textbf{48.4} & 45.8 & \textbf{48.0} \\ \hline
			\multirow{5}{*}{10} & YOLO-FS~\cite{yolore} & 30.0 & 62.7 & 43.2 & 60.6 & 39.6 & 47.2 & 43.2 & 13.9 & 41.5 & 58.1 & 39.2 & 39.2 & 20.1 & 51.8 & 55.6 & 42.4 & 36.6 & 41.3 \\
			& Meta R-CNN~\cite{metarcnn} & \textbf{52.5} & 55.9 & 52.7 & 54.6 & 41.6 & 51.5 & \textbf{52.8} & 3.0 & 52.1 & \textbf{70.0} & 49.2 & 45.4 & 13.9 & \textbf{72.6} & 58.3 & 47.8 & 47.6 & 48.1 \\
			& Baseline & 38.6 & 48.6 & 51.6 & 57.2 & 43.4 & 47.9 & 46.9 & 14.8 & 42.1 & 57.4 & 43.4 & 40.9 & 18.2 & 59.1 & 57.3 & 50.1 & 41.5 & 45.2 \\
			& Baseline-FPN & 41.8 & 68.4 & 61.7 & 66.8 & 45.8 & 56.9 & 52.7 & 16.3 & 46.8 & 58.1 & 44.9 & 43.8 & \textbf{25.8} & 50.2 & \textbf{67.7} & 47.7 & 47.8 & 47.8 \\
			& MPSR (ours) & 48.3 & \textbf{73.7} & \textbf{68.2} & \textbf{70.8} & \textbf{48.2} & \textbf{61.8} & 51.8 & \textbf{16.7} & \textbf{53.1} & 66.4 & \textbf{51.2} & \textbf{47.8} & 24.4 & 55.8 & 67.5 & \textbf{50.4} & \textbf{50.5} & \textbf{49.7} \\ \hline
		\end{tabular}
	\end{center}
\end{sidewaystable}
\setlength{\tabcolsep}{1.4pt}
\end{document}